\documentclass{article}

\usepackage[nonatbib, preprint]{nips_2018}

\usepackage[utf8]{inputenc} 
\usepackage[T1]{fontenc}    
\usepackage{hyperref}       
\usepackage{url}            
\usepackage{booktabs}       
\usepackage{amsfonts}       
\usepackage{nicefrac}       
\usepackage{microtype}      

\usepackage{amsfonts}       

\usepackage{forloop}
\usepackage{mathtools}

\usepackage{algorithm}
\usepackage{algpseudocode}
\usepackage{graphicx}
\usepackage{xcolor}
\graphicspath{{figure/}}

\definecolor{c:recon}{rgb}{0.15,0.35,0.5}
\definecolor{c:ent}{rgb}{0.1,0.5,0.2}
\definecolor{c:dyn}{rgb}{0.2,0,0.4}
\definecolor{c:forward}{rgb}{0,0.44,0.73}
\definecolor{c:backward}{rgb}{0.75, 0.15, 0.18}

\newcommand{\defvec}[1]{\expandafter\newcommand\csname v#1\endcsname{{\mathbf{#1}}}}
\newcounter{ct}
\forLoop{1}{26}{ct}{
    \edef\letter{\alph{ct}}
    \expandafter\defvec\letter
}
\forLoop{1}{26}{ct}{
    \edef\letter{\Alph{ct}}
    \expandafter\defvec\letter
}

\DeclareMathOperator{\grad}{\nabla}
\DeclareMathOperator{\KL}{\mathbb{D}_{\text{KL}}}
\DeclareMathOperator{\SE}{\mathbb{H}}
\DeclareMathOperator{\E}{\mathbb{E}}

\DeclareMathOperator{\vecOp}{\rm vec}

\newcommand{\half}{\frac{1}{2}}
\newcommand{\bm}[1]{\boldsymbol{\mathbf{#1}}}

\allowdisplaybreaks

\title{Variational online learning of neural dynamics}

\author{
	Yuan Zhao \\
	Department of Neurobiology and Behavior \\
	Stony Brook University \\
	Stony Brook, NY 11794 \\
	\texttt{yuan.zhao@stonybrook.edu} \\
    \And
    Il Memming Park \\
    Department of Neurobiology and Behavior \\
    Department of Applied Mathematics and Statistics \\
    Institute for Advanced Computational Science \\
	Stony Brook University \\
    Stony Brook, NY 11794 \\
    \texttt{memming.park@stonybrook.edu} \\
}

\begin{document}


\maketitle

\begin{abstract}
	New technologies for recording the activity of large neural populations during complex behavior provide exciting opportunities for investigating the neural computations that underlie perception, cognition, and decision-making.
	Nonlinear state space models provide an interpretable signal processing framework by combining an intuitive dynamical system with a probabilistic observation model, which can provide insights into neural dynamics, neural computation, and development of neural prosthetics and treatment through feedback control.
	It brings the challenge of learning both latent neural state and the underlying dynamical system because neither is known for neural systems \textit{a priori}.
	We developed a flexible online learning framework for latent nonlinear state dynamics and filtered latent states. Using the stochastic gradient variational Bayes approach, our method jointly optimizes the parameters of the nonlinear dynamical system, the observation model, and the black-box recognition model.
	Unlike previous approaches, our framework can incorporate non-trivial distributions of observation noise and has constant time and space complexity. These features make our approach amenable to real-time applications and the potential to automate analysis and experimental design in ways that testably track and modify behavior using stimuli designed to influence learning.
\end{abstract}

\section{Introduction}

Discovering interpretable structure from a streaming high-dimensional time series has many applications in science and engineering.
Since the invention of the celebrated Kalman filter, state space models have been successful in providing a succinct, hence more interpretable, description of the underlying dynamics that explains the observed time series as trajectories in a low-dimensional state space.
Taking a step further, state space models equipped with nonlinear dynamics provide an opportunity to describe the latent ``laws'' of the system that is generating the seemingly entangled time series~\cite{Haykin1998,Ko2009,Mattos2016}.
Specifically, we are concerned with the problem of identifying a continuous nonlinear dynamics in the state space $\vx(t) \in \mathbb{R}^d$ that captures the spatiotemporal structure of a noisy observation $\vy(t)$:
\begin{subequations}\label{eq:continuous:ssm}
	\begin{align}
		\dot{\vx} & = F_\theta(\vx(t), \vu(t))
		\,        & \text{(state dynamics)}
		\label{eq:sde}
		\\
		\vy(t)    & \sim P(\vy(t) \mid G_\theta(\vx(t), \vu(t))) \, & \text{(observation model)}
		\label{eq:obs}
	\end{align}
\end{subequations}
where $F$ and $G$ are continuous functions that may depend on parameter $\theta$, $\vu(t)$ is control input, and $P$ denotes a probability distribution that captures the noise in the observation; e.g., Gaussian distribution for field potentials or Poisson distribution for spike counts.

In practice, the continuous-time state dynamics is more conveniently formulated in discrete time as,
\begin{align}\label{eq:dynamics}
	\vx_{t+1} & = f_\theta(\vx_t, \vu_t) + \bm{\epsilon}_t \, & \text{(discrete time state dynamics)}
\end{align}
where $\bm{\epsilon}_t$ is intended to capture the unobserved (latent) perturbations of the state $\vx_t$.
Such (spatially) continuous state space models are natural in many applications where the changes are slow and the underlying system follows physical laws and constraints (e.g., object tracking), or where learning the laws are of great interest (e.g. in neuroscience and robotics)~\cite{Roweis2001,Mante2013,Sussillo2013,Frigola2014,Zhao2017}.
Specifically, in the context of neuroscience, the state vector $\vx_t$ represents the instantaneous state of the neural population, while $f$ captures the time evolution of the population state.
Further interpretation of $f$ can provide understanding as to how neural computation is implemented~\cite{Zhao2016,Mante2013,Russo2018}.

If the nonlinear state space model is fully specified, Bayesian inference methods can be employed to estimate the current state~\cite{Ho1964,Sarkka2013}.
Conventionally, the estimation of latent states using only the past observation is referred to as filtering -- inference of the filtering distribution, $p(\vx_t \mid \vy_{\le t})$.
If both past and future observations are used, then the quantity of interest is usually the smoothing distribution, $p(\vx_{\le t} \mid \vy_{\le t})$).
We are also interested in predicting the distribution over future states, $p(\vx_{t:t+s} \mid \vy_{\le t})$ and observations, $p(\vy_{t+1:t+s} \mid \vy_{\le t})$ for $s > 0$.
However, in many applications, the challenge is in learning the parameters $\theta$ of the state space model (a.k.a. the system identification problem).
We aim to provide a method for simultaneously learning both the latent trajectory $\vx_t$ and the latent (nonlinear) dynamical and observational system $\theta$, known as the \textit{joint estimation problem}~\cite{Haykin2001}.

Expectation maximization (EM) based methods have been widely used in practice~\cite{Ghahramani1999,Valpola2002,Turner2010,Golub2013}, and more recently variational autoencoder methods~\cite{Archer2015,Krishnan2015,Johnson2016,Krishnan2017, Karl2017,Watter2015} have been proposed, all of which are designed for offline analysis, and not appropriate for real-time applications.
Recursive stochastic variational inference has been successful in streaming data assuming independent samples~\cite{Broderick2013}, however, in the presence of temporal dependence, proposed variational algorithms (e.g.~\cite{Frigola2014}) remain theoretical and lack testing.

In this study, we are interested in \textit{real-time} signal processing and state space control setting~\cite{Golub2013} where online algorithms are needed that can recursively solve the joint estimation problem on streaming observations.
A popular solution to this problem exploits the fact that online state estimators for nonlinear state space models such as extended Kalman filter (EKF) or unscented Kalman filter (UKF) can be used for nonlinear regression formulated as a state space model.
By augmenting the state space with the parameters, one can build an online \textit{dual} estimator using nonlinear Kalman filtering~\cite{Wan2000,Wan2001}.
However, they involve coarse approximation of Bayesian filtering, involve many hyperparameters, do not take advantage of modern stochastic gradient optimization, and are not easily applicable to arbitrary observation likelihoods.
There are also closely related online version of EM-type algorithms~\cite{Roweis2001} that share similar concerns.

In hopes of lifting these concerns, we derive an \textit{online} black-box variational inference framework, referred to as \textbf{variational joint filtering (VJF)}, applicable to a wide range of nonlinear state dynamics (dynamic models) and observation models, that is, the computational demand of the algorithm is constant per time step.
Our approach aims at
\begin{enumerate}
	\item \textbf{Online adaptive learning}: Our target application scenarios are streaming data. This allows the inference during an experiment or as part of a neural prosthetics. If the system changes, the inference will catch up with the altered system parameters.
	\item \textbf{Joint estimation}: The proposed method is supposed to simultaneously learn the latent states $p(\vx_t \mid \vy_{\le t})$, state dynamics $f(\vx_t, \vu_t)$ and the observation model $G(\vx, \vu)$. No offline training is necessary to learn the system parameters.
	\item \textbf{Interpretability}: Under the framework of state space modeling, rather than interpret the system via model parameters, we employ the language of dynamical systems and capture the characteristics of the system qualitatively via fixed point, limit cycle, strange attractor, bifurcation and so on which are key components of theories of neural dynamics and computation.
\end{enumerate}
We focus on low-dimensional latent dynamics that often underlie many neuroscientific experiments and allow for producing interpretable visualizations of complex collective network dynamics in this study.

\section{Variational Principle for Online Joint Estimation}

The crux of recursive Bayesian filtering is updating the posterior over the latent state one step at a time:
\begin{align}\label{eq:update:exact}
	p( \vx_t \mid \vy_{\le t} ) & =
	\underbrace{p( \vy_t \mid \vx_t )}_{\substack{\text{likelihood}}} \,
	\underbrace{p( \vx_t \mid \vy_{<t} )}_{\substack{\text{prior at time $t$}}}
	\;
	/\!\!\!
	\underbrace{p(\vy_t \mid \vy_{<t})}_{\substack{\text{marginal likelihood}}}
\end{align}
where the input $\vu_t$ and parameters $\theta$ are omitted for brevity.
Unfortunately, the exact calculations of Eq.~\eqref{eq:update:exact} are not tractable in general, especially for nonlinear dynamic models and/or non-conjugate distributions.
We thus turn to approximate inference and develop a recursive variational Bayesian filter by deriving an evidence lower bound for the marginal likelihood as the objective function.
Let $q(\vx_t)$ denote an arbitrary probability measure which will eventually approximate the filtering density $p(\vx_t \mid \vy_{\le t})$.
From Eq.~\eqref{eq:update:exact}, we can rearrange the marginal log-likelihood as
\begin{equation}
	\begin{aligned}
		 & \log p(\vy_{t} \mid \vy_{<t})                                                                                                                                                           \\
		 & = \log \frac{p(\vy_{t} \mid \vx_{t}) p(\vx_{t} \mid \vy_{<t})}{p(\vx_{t} \mid \vy_{\le t})} \qquad \text{for any $\vx_t$} \nonumber
		\\
		 & = \E_{q(\vx_{t})} \left[\log \frac{p(\vy_{t} \mid \vx_{t}) p(\vx_{t} \mid \vy_{<t})}{p(\vx_{t} \mid \vy_{\le t})}\right] \qquad \text{the marginal is constant to $q(\vx_t)$} \nonumber
		\\
		 & = \E_{q(\vx_{t})} \left[
			\log \frac{p(\vy_{t} \mid \vx_{t}) p(\vx_{t} \mid \vy_{<t}) q(\vx_{t})}{p(\vx_{t} \mid \vy_{\le t}) q(\vx_{t})}
			\right]
		\\
		 & =
		{\color{c:recon}\underbrace{\E_{q(\vx_{t})} \left[ \log p(\vy_{t} \mid \vx_{t}) \right]}_{\substack{\text{reconstruction log-likelihood}}}}
		- \KL(q(\vx_{t}) \,\|\, p(\vx_{t} \mid \vy_{<t}))
		+ \underbrace{\KL(q(\vx_{t}) \,\|\, p(\vx_{t} \mid \vy_{\le t}))}_{\substack{\color{red} \text{variational gap \#1}}}
		\\
		 & \ge {\color{c:recon}\E_{q(\vx_{t})} \left[
				\log p(\vy_{t} \mid \vx_{t})
				\right]}
		+ {\color{c:ent}\underbrace{\SE(q(\vx_{t}))}_{\substack{\text{entropy}}}}
		+ \E_{q(\vx_{t})} \left[
			\log p(\vx_{t} \mid \vy_{<t}) \right]
		\\
		 & = {\color{c:recon} \E_{q(\vx_{t})} \left[ \log p(\vy_{t} \mid \vx_{t}) \right]}
		+ {\color{c:ent} \SE(q(\vx_{t}))}
		+ \E_{q(\vx_{t})} \left[
			\log
			\E_{p( \vx_{t-1} \mid \vy_{<t})} \left[
				p(\vx_t \mid \vx_{t-1})
				\right]
			\right]
		\\
		 & = {\color{c:recon} \E_{q(\vx_{t})} \left[ \log p(\vy_{t} \mid \vx_{t}) \right]}
		+ {\color{c:ent} \SE(q(\vx_{t}))}
		+ \E_{q(\vx_{t})} \left[
			\E_{p(\vx_{t-1} | \vy_{<t})}
			\left[
				\log p(\vx_{t} \mid \vx_{t-1})
				\right]\right]\nonumber
		\\
		 & \quad +
		\underbrace{
			\E_{q(\vx_{t})} \left[
				\KL(p(\vx_{t-1}\mid \vy_{<t})\,\|\,p(\vx_{t-1} \mid \vx_{t},\vy_{<t}))
				\right]
		}_{\substack{\color{red} \text{variational gap \#2}}}                                                                                                                                      \\
		 & \ge
		{\color{c:recon} \E_{q(\vx_{t})} \left[ \log p(\vy_{t} \mid \vx_{t}) \right]}
		+ {\color{c:ent} \SE(q(\vx_{t}))}
		+ \E_{q(\vx_{t})}
		\E_{ p( \vx_{t-1} \mid \vy_{<t}) }
		\left[
			\log p(\vx_t \mid \vx_{t-1})
			\right]
	\end{aligned}
\end{equation}
where $\SE$ denotes Shannon's entropy and $\KL$ denotes the Kullback-Leibler (KL) divergence~\cite{Cover1991}.
Maximizing this lower bound would result in a variational posterior
$q(\vx_t) \approx p(\vx_t \mid \vy_{\le t})$ w.r.t. $q(\vx_t)$.
Naturally we plug in the previous step's solution to the next time step, obtaining a loss function suitable for recursive estimation:
\begin{equation} \label{eq:lbound}
	\begin{aligned}
		\mathcal{L}
		\coloneqq &
		{\color{c:recon} \E_{q(\vx_{t})} \left[ \log p(\vy_{t} \mid \vx_{t}) \right]}
		+ {\color{c:ent} \SE(q(\vx_{t}))}
		+ {\color{c:dyn} \underbrace{
			\E_{q(\vx_{t})}
			\E_{\color{red} q( \vx_{t-1} ) }
			\left[
				\log p(\vx_t \mid \vx_{t-1})
				\right]
		}_{\substack{\text{dynamics log-likelihood}}}}
	\end{aligned}
\end{equation}
This also results in consistent $q(\vx_t)$ for all time steps as they are in the same family of distribution.

Meanwhile, as it is aimed to jointly estimate the observation model $p(\vy_t \mid \vx_t)$ and state dynamics $p(\vx_t \mid \vx_{t-1})$, we achieve online inference by maximizing this objective $\mathcal{L}$ w.r.t. their parameters (omitted for brevity) and the variational posterior distribution $q(\vx_t)$ simultaneously provided that $q(\vx_{t-1})$ takes some parameterized form and has been estimated from the previous time step.
Maximizing the objective $\mathcal{L}$ is equivalent to minimizing the two variational gaps:
(1) the variational filtering posterior has to be close to the true filtering posterior, and
(2) the filtering posterior from the previous step needs to be close to $p(\vx_{t-1}\mid \vx_t, \vy_{<t})$.
Note that the second gap is invariant to $q(\vx_t)$ if $p(\vx_{t-1}\mid \vx_t, \vy_{<t}) = p(\vx_{t-1} \mid \vy_{<t})$, that is, the one-step backward smoothing distribution is identical to the filtering distribution.

On the flip side, intuitively, there are three components in $\mathcal{L}$ that are jointly optimized: (1) \textcolor{c:recon}{reconstruction log-likelihood} which is maximized if $q(\vx_t)$ concentrates around the maximum likelihood estimate given only $\vy_t$, (2) the \textcolor{c:dyn}{dynamics log-likelihood} which is maximized if $q(\vx_t)$ concentrates at around the maximum of $\E_{q(\vx_{t-1})}\left[ \log p(\vx_t\mid\vx_{t-1}) \right]$, and (3) the \textcolor{c:ent}{entropy} that expands $q(\vx_t)$ and prevents it from collapsing to a point mass.

In order for this recursive estimation to be real-time, we choose $q(\vx_t)$ to be a multivariate normal with diagonal covariance $\mathcal{N}(\bm{\mu}_t, \vs_{t})$ where $\bm{\mu}_t$ is the mean vector and $\vs_{t}$ is the diagonal of the covariance matrix in this study.
Moreover, to amortize the computational cost of optimization to obtain the best $q(\vx_t)$ on each time step, we employ the variational autoencoder architecture~\cite{Hinton1995} to parameterize $q(\vx_t)$ with a recognition model.
Intuitively, the recognition model embodies the optimization process of finding $q(\vx_t)$, that is, it performs an approximate Bayesian filtering computation (in constant time) of Eq.~\eqref{eq:update:exact} according to the objective function $\mathcal{L}$.
We use a recursive recognition model that maps $q(\vx_{t-1})$ and $\vy_t$ to $q(\vx_{t})$.
In particular, the recognition model takes a deterministic recursive form:
\begin{align}\label{eq:recog}
	[\bm{\mu}_{t}, \vs_{t}] =
	h(\vy_{t}, \bm{u}_{t-1}, \bm{\mu}_{t-1}, \vs_{t-1})
\end{align}
Specifically $h$ takes a simple the form of the multi-layer perceptron (MLP)~\cite{Hastie2009} in this study, and we refer to its parameters as the recognition model parameters.
Note that the recursive architecture of the recognition model reflects the Markovian structure of the assumed dynamics (c.f., smoothing networks often use bidirectional recurrent neural network (RNN)~\cite{Sussillo2016} or graphical models~\cite{Archer2015,Johnson2016}).

The expectations appearing in the \textcolor{c:recon}{reconstruction log-likelihood} and \textcolor{c:dyn}{dynamics log-likelihood} are not always tractable in general. For those intractable cases, one can use the reparameterization trick and stochastic variational Bayes~\cite{Rezende2014,Kingma2014}: rewriting the expectations over $q$ as expectation over a standard normal random variable, i.e., $\bm{\mu}_t + \vs_t^{\frac{1}{2}} \mathcal{N}(0, 1)$, and using a single sample for each time step.
Hence, in practice, we optimize the following objective function (the other variables and parameters are omitted for brevity),
\begin{equation}\label{eq:lbound_reparam}
	\hat{\mathcal{L}} =
	\log p(\vy_{t} \mid \tilde\vx_{t}, \theta) + \E_{q(\vx_{t})}\log p(\vx_t \mid \tilde\vx_{t-1}, \theta)
	+ H(q(\vx_{t})) 
\end{equation}
where $\tilde\vx_t$ and $\tilde\vx_{t-1}$ represent random samples from $q(\vx_t)$ and $q(\vx_{t-1})$ respectively.
Note that the remaining expectation over $q(\vx_t)$ has closed form solution under our Gaussian state noise, $\bm{\epsilon}_t$, assumption.
Thus, our method can handle arbitrary observation and dynamic model unlike dual form nonlinear Kalman filtering methods that usually suffer difficulties in sampling, e.g. transforming Gaussian random numbers into point process observations.

\begin{figure}[!ht]
	\centering
	\includegraphics[width=0.75\textwidth]{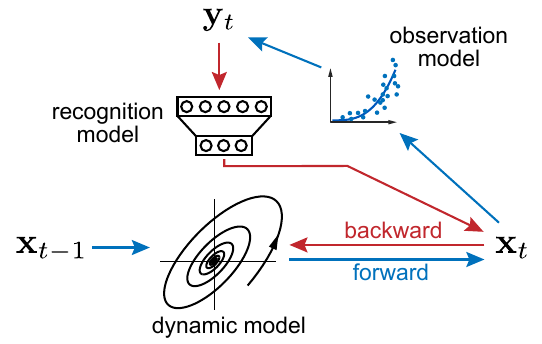}
	\caption[Diagram]{Schematics of variational joint filtering.
		\textcolor{c:forward}{Blue} arrows indicates forward pass in which the previous latent state generates the new state through the state dynamics, and the new state transforms into the observation through the observation model. \textcolor{c:backward}{Red} arrows indicate backward pass in which the recognition model recovers the current latent state from the observation. The three components, observation model, recognition model and dynamical system, are updated by backpropagation.}
	\label{fig:diagram}
\end{figure}

The schematics of the proposed inference algorithm is summarized by two passes in Figure~\ref{fig:diagram}.
In the \textcolor{c:forward}{forward pass}, the previous latent state generates the new state through the dynamic model, and the new state transforms into the observation through the observation model. In the \textcolor{c:backward}{backward pass}, the recognition model recovers the current latent state from the observation, and the observation model, recognition model and dynamic model, are updated by backpropagation.

Algorithm~\ref{alg:filtering} is an overview of the recursive estimation algorithm.
Denoting the set of all parameters by $\bm{\Theta}$ of the observation model, recognition model and dynamic models, the objective function in Eq.~\eqref{eq:lbound_reparam} is differentiable w.r.t. $\bm{\Theta}$, and thus we employ empirical Bayes and optimize it through stochastic gradient ascent (using Adam~\cite{Kingma2014a}).
We outline the algorithm for a single vector time series, but we can filter multiple time series with a common state space model simultaneously, in which case the gradients are averaged across the instantiations.
Note that this algorithm has \textit{constant time and space complexity} per time step.

In practice, the measurements $\vy_t$ and input $\bm{u}_t$ are sampled at a regular interval.
Algorithm~\ref{alg:filtering} is called after every such observation event, which will return the state estimate along with the parameters and the dynamical system.
One can visualize these for real-time for monitoring, and/or have it streamed to another system for further automated processing (e.g. detect anomalies and raise an alarm or deliver feedback controls).

\begin{algorithm}[!ht]
	\caption{Variational Joint Filtering (single step)}
	\label{alg:filtering}
	\begin{algorithmic}
		\Procedure{VJF}{$\vy_{t}, \bm{u}_{t-1}, \bm{\mu}_{t-1}, \vs_{t-1}, \bm{\Theta}$}
		\State $\epsilon_{t} \gets \mathcal{N}(\bm{0}, \vI), \; \epsilon_{t-1} \gets \mathcal{N}(\bm{0}, \vI)$ \Comment{Draw random samples}
		\State [$\bm{\mu}_{t}, \vs_{t}] \coloneqq \vh(\vy_{t}, \vu_{t-1}, \bm{\mu}_{t-1}, \vs_{t-1})$ \Comment{State estimation}
			\State $\tilde\vx_{t} \coloneqq \bm{\mu}_{t} + \vs_{t}^{\nicefrac{1}{2}} \epsilon_t$
			\State $\tilde\vx_{t-1} \coloneqq \bm{\mu}_{t-1} + \vs_{t-1}^{\nicefrac{1}{2}} \epsilon_{t-1}$ %
			\State Update $\bm{\Theta}$ with $\grad_{\bm{\Theta}}
		\hat{\mathcal{L}}(\bm{\Theta}; \vy_{t}, \tilde\vx_{t}, \tilde\vx_{t-1}, \bm{u}_{t-1})$
			\Comment{Model update}
			\State\Return $\bm{\mu}_{t}, \vs_{t}$ and $\bm{\Theta}$
		\EndProcedure
	\end{algorithmic}
\end{algorithm}

\section{Application to Latent Neural Dynamics}

Our primary applied aim is real-time neural interfaces where a population of neurons are recorded while a low-dimensional stimulation is delivered~\cite{Newman2015,ElHady2016,Hocker2019a}.
State-space modeling of such neural time series have been successful in describing population dynamics~\cite{Macke2011, Zhao2017}.
Moreover, models of neural computation are often described as dynamical systems~\cite{Hopfield1982,Dayan2001,Barak2013}.
For example, attractor dynamics where the convergence to one of the attractors represents the result of computation~\cite{Wang2002,Nassar2018b}.
Here we propose a parameterization and tools for visualization of the model suitable for studying neural dynamics and building neural interfaces~\cite{Zhao2016}.
In this section, we provide methodological details for the results presented in the next section.

\subsection{Parameterization of the state space model}

Having in mind high-temporal resolution neural spike trains where each time bin has at most one action potential per channel, we describe the case for point process observation.
However, note that our method generalizes to arbitrary observation likelihoods that are appropriate for other modalities, including calcium imaging or local field potentials.
The observed point process time series $\vy_t$ is a stream of sparse binary vectors.
All experiments of point process observation were binned finely so that the time bins contain one event each at most in this study.

Our observation model, Eq.~\eqref{eq:lik}, assumes that the observation vector $\vy_{t}$ is sampled from a probability distribution $P$ determined by the latent state $\vx_{t}$ though a linear-nonlinear map possibly together with extra parameters at each time $t$,
\begin{equation}
	\label{eq:lik}
	\vy_t \sim P(g(\vC \vx_t + \vb))
\end{equation}
where $g\colon \mathbb{R} \to \mathbb{R}$ is a point-wise map.
We use the canonical link $g(\cdot) = \exp(\cdot)$ for Poisson likelihood and identity for Gaussian likelihood in this study.
Note that this observation model is not identifiable since $\vC\vx_{t} = (\vC\vR) (\vR^{-1}\vx_{t})$ where $\vR$ is an arbitrary invertible matrix.
We normalize the loading matrix $\vC$ in each iteration.
It is straightforward to include more additive exogenous variables, history-filter for refractory period, coupling between processes, and stimulation artifacts~\cite{Pillow2008,Truccolo2005}.

For state dynamic model, we propose to use a specific additive parameterization with  state transition function and input interaction as a special case of Eq.~\eqref{eq:dynamics},
\begin{subequations} \label{eq:generative}
	\begin{align}
		\vx_{t+1}                & = \vx_{t} + f(\vx_{t}) + \vB_t \vu_{t} + \bm{\epsilon}_{t+1} \\
		f(\vx_t)                 & = \vW \bm{\phi}({\vx}_t)                                     \\
		\vx_{0}, \bm{\epsilon}_t & \sim \mathcal{N}(\bm{0}, \sigma^2 \vI) \label{eq:prior:n}
	\end{align}
\end{subequations}
where $\bm{\phi}(\cdot)$ is a vector of $r$ continuous basis functions, i.e.
$\bm{\phi}(\cdot) = (\phi_1(\cdot), \ldots, \phi_r(\cdot))^\top$, $\vW$ is the weight matrix of the radial basis functions, and $\vB_t$ is the interaction with the input $\vu_t$. The interaction $\vB_t$ can be globally linear, parameterized as a matrix independent from $\vx_t$, or locally linear, parameterized as a matrix-valued function of $\vx_t$ using also RBF networks. i.e. $\vecOp(\vB(\vx_t)) = \vW_B \bm{\phi}({\vx}_t)$ where $\vW_B$ is the respective weight matrix.
In this study, we used squared exponential radial basis functions~\cite{Roweis2001,Frigola2014,Sussillo2013,Zhao2016},
\begin{equation}
	\phi_i(\vx) = \exp\left(-\half \gamma_i \lVert \vx - \vc_i \rVert_2^2\right)
\end{equation}
with centers $\vc_i$ and corresponding inverse squared kernel width $\gamma_i$.
Though the dynamics can be modeled by other universal approximators such as percepton and RNN, we chose the radial basis function network for the reasons of non-wild extrapolation (zero velocity when the state is far away from data) and fast computation.


The time complexity of our algorithm is $\mathcal{O}(mpr + n(m + p + q))$ where $n, m, p, q, r$ denote the dimensions of observation, latent space, input, the numbers of hidden units and radial basis functions for this specific parameterization.
Practically to achieve realistic computation time for real-time applications in neuroscience, the number of radial basis functions and hidden units are constrained by the requirement.
Note that the time complexity does not grow with time that enable efficient online inference.
If we compare this to an efficient offline algorithm such as PLDS~\cite{Macke2011} run repeatedly for every new observation (``online mode''), its time complexity is $\mathcal{O}(t \cdot (m^3 + mn))$ per time step at time $t$ which grows as time passes, making it impractical for real-time application.

\subsection{Phase portrait analysis}
Phase portrait displays key qualitative features of dynamics, and with a little bit of training, it provides a visual means to interpreting dynamical systems.
The law that governs neural population dynamics captured in the inferred function $f(\vx)$ directly represents the velocity field of an underlying smooth dynamics \eqref{eq:sde} in the absence of input~\cite{Roweis2001,Zhao2016}.
In the next section, we visualize the estimated dynamics as phase portrait which consists of the vector field, example trajectories, and estimated dynamical features (namely fixed points)~\cite{Strogatz2000}.
We can numerically identify candidate fixed points $\vx^\ast$ that satisfy $f(\vx^\ast) \approx 0$.
For the synthetic experiments, we performed an affine transformation to orient the phase portrait to match the canonical equations in the main text when the simulation is done through the proposed observation model if the observation model is unknown and estimated.

\vspace{1ex}
\subsection{Prediction}
For state space models, we can predict both future latent trajectory and future observations.
The $s$-step ahead prediction can be sampled from the predictive distributions:
\begin{subequations}
	\begin{align}
		p(\vx_{t+1:t+s} \mid \vy_{\le t}) = & \mathbb{E}_{q(\vx_{t})} [p(\vx_{t+1:t+s} \mid \vx_{t})]
		\\
		p(\vy_{t+1:t+s} \mid \vy_{\le t}) = & \mathbb{E}_{p(\vx_{t+1:t+s} \mid \vy_{\le t})} [p(\vy_{t+1:t+s} \mid \vx_{t+1:t+s})]
	\end{align}
\end{subequations}
given estimated parameters by current time $t$ without seeing the data $\vy_{t+1:t+s}$ during these steps.
In the figures of experiments, we plot the mean of the predictive distribution as trajectories.

\section{Experiments on Theoretical Models of Neural Computation}

We demonstrate our method on a range of nonlinear dynamical systems relevant to neuroscience.
Many theoretical models have been proposed in neuroscience to represent different schemes of computation.
For the purpose of interpretable visualization, we choose to simulate from two or three dimensional dynamical systems. We apply the proposed method to four such low-dimensional models: a ring attractor model as a model of internal head direction representation, an nonlinear oscillator as a model of rhythmic population-wide activity, a biophysically realistic cortical network model for a visual discrimination experiment and a chaotic attractor.

In the synthetic experiments, we first simulated state trajectories by respective differential equations, and then generated either Gaussian or point process observations (to mimic spikes) via Eq.~\eqref{eq:lik} with corresponding distributions. The parameters $\vC$ and $\vb$ were randomly drawn, and they were constrained to keep firing rate $<60$~Hz on average for realistic spiking behavior. All observations are spatially $200$-dimensional unless otherwise mentioned.
We refer to their conventional formulations under different coordinate systems, but our simulations and inferences are all done in Cartesian coordinates.
Note that we focus on online learning in this study and always train our model with streaming data, even while comparing with offline methods.

The approximate posterior distribution is defined recursively in Eq.~\eqref{eq:recog} as diagonal Gaussian with mean and variance determined by corresponding observation, input and previous step via a recurrent neural network.
We used a one-hidden-layer MLP in this study. Typically the state noise variance $\sigma^2$ is unknown and has to be estimated from data. To be consistent with Eq.~\eqref{eq:prior:n}, we set the starting value of $\sigma^2$ to be $1$, and hence $\bm{\mu}_{0} = \bm{0}, \vs_{0} = \vI$.
We initialize the loading matrix $\vC$ by factor analysis, and column-wisely normalize it by $\ell_2$ norm every iteration to keep the system identifiable.

\vspace{1ex}
\subsection{Ring attractor}

Continuous attractors are often used as models for neural representation of continuous variables~\cite{Sussillo2013, Mante2013}. For example, a bump attractor network with ring topology is proposed as the dynamics underlying the persistently active set of neurons that are tuned for the angle of the animal's head direction~\cite{Peyrache2015}. Here we use the following 2-variable reduction of the ring attractor system:
First, we study the following two-variable ring attractor system:
\begin{equation}\label{eq:ring}
	\begin{aligned}
		\tau_r \dot{r}             & = r_0 - r \\
		\tau_\varphi \dot{\varphi} & = I
	\end{aligned}
\end{equation}
where $\varphi$ represents the direction driven by input $I$, and $r$ is the radial component representing an internal circular variable, such as head direction.
We simulated 100 trajectories (1000 steps) with step size $\Delta t = 0.1$, $r_0=1$, $\tau_r = 1$, $\tau_\varphi=1$ with Gaussian state noise ($\text{std}=0.005$) added each step.
Though the ring attractor is defined in polar coordinate system, we transformed it into Cartesian system for simulation and training. In simulation we used strong input (tangent drift) to keep the trajectories flowing around the ring clockwise or counter-clockwise. The point process observations were generated by passing the states through a linear-nonlinear map (Eq.~\eqref{eq:lik}) and sampling from a Poisson distribution.
We streamed the observations into the proposed algorithm that consists of point process likelihood, dynamic model with 20 radial basis functions and locally linear input interaction in Eq.~\eqref{eq:dynamics} and a recognition MLP with 100 hidden units.

\begin{figure}[!ht]
	\centering
	\includegraphics[width=\textwidth]{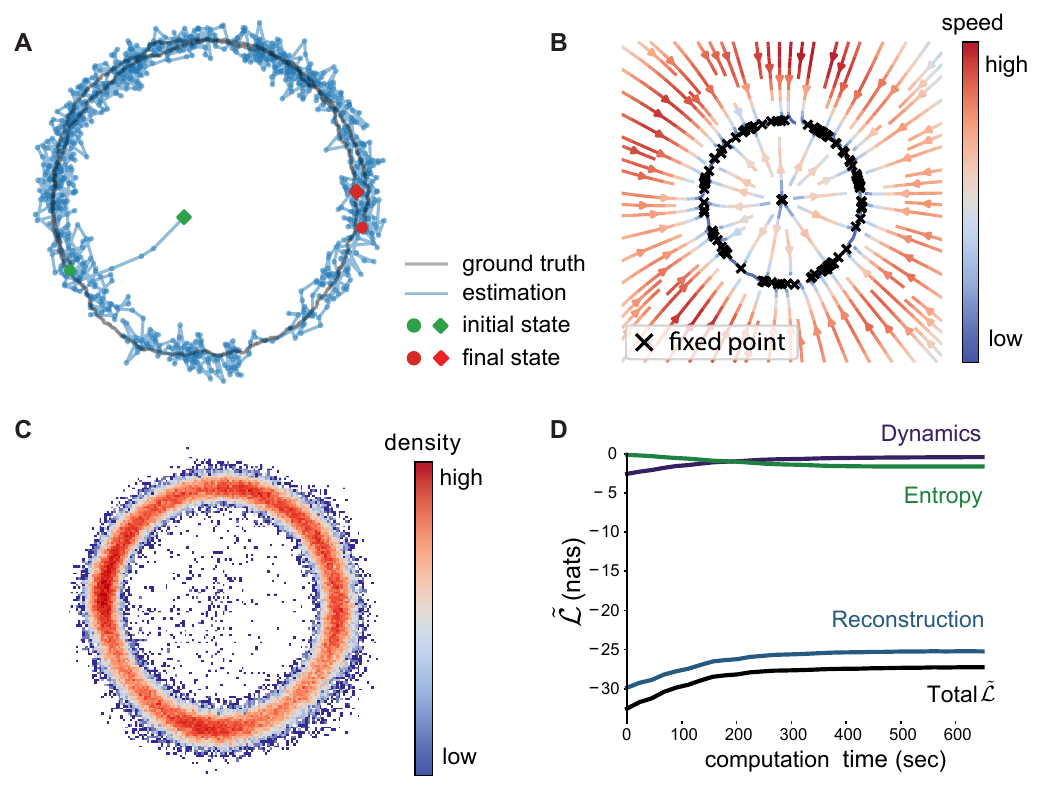}
	\caption[Ring attractor model]{
		Ring attractor model.
		\textbf{(A)} One latent trajectory (black) in the training set and the corresponding filtered mean $\mu_t$ (blue).
		\textbf{(B)} Velocity field reconstructed from the trained proposed model. The colored streamlines indicates the speed and the directions. The black crosses are candidate fixed points obtained from inferred dynamics. Note the collection of fixed points around the ring shape. The central fixed point is unstable.
		\textbf{(C)}Density of the posterior means. The density of inferred means of all trajectories in the training set. The higher it is, the more confidence we have on the inferred dynamics where we have more data.
		\textbf{(D)} Convergence on the ring attractor.
		We display the three components of the objective lower bound: reconstruction log-likelihood, dynamics log-likelihood, entropy, and the lower bound itself from Eq.~\eqref{eq:lbound}.
		The average computation time per step is 1.1~ms (more than 900 data points per sec).
	}
	\label{fig:ring}
\end{figure}

Figure~\ref{fig:ring}A illustrates one latent trajectory (black) and its variational posterior mean (blue).
These two trajectories start at green circle and diamond respectively and end at the red markers. The inference starts near the center (origin) that is relatively far from the true location because the initial posterior mean is set at zero. The final states are very close which implies that the recognition model works well. Figure~\ref{fig:ring}B shows the reconstructed velocity field by the model.
We visualized the velocity as colored directional streamlines. We can see the velocity toward the ring attractor and the speed is smaller closer to the ring.
The model also identifies a number of fixed points arranged around the ring attractor via numerical roots finding.
Figure~\ref{fig:ring}C shows the distribution of posterior means of all data points in the state space. We have more confidence of the inferred dynamical system in the denser area.

Figure~\ref{fig:ring}D shows the three components of Eq.~\eqref{eq:lbound} and the objective lower bound clearly, demonstrating the convergence of the algorithm.
We can see each component reaches a plateau within $400$ sec.
As the reconstruction and dynamics log-likelihoods increase, the recognition model and dynamical model are getting more accurate while the decreasing entropy indicates the increasing confidence (inverse posterior variance) on the inferred latent states. The average computation time of a joint estimation step is $1.1$~ms
(hardware specification: Intel Xeon E5-2680 2.50G~Hz, 128GB RAM, no GPU).

\vspace{1ex}
\subsection{Nonlinear oscillator}

Dynamical systems have been a successful application in the biophysical models of single neuron in neuroscience. We used a relaxation oscillator, the FitzHugh-Nagumo (FHN) model~\cite{Izhikevich2007}, which is a 2-dimensional reduction of the Hodgkin-Huxley model:
We used a  with the following nonlinear state dynamics:
\begin{equation}
	\label{eq:FHN}
	\begin{aligned}
		\dot{v} & = v(a - v)(v - 1) - w + I, \\
		\dot{w} & = bv - cw,
	\end{aligned}
\end{equation}
where $v$ is the membrane potential, $w$ is a recovery variable and $I$ is the magnitude of stimulus current in modeling single neuron biophysics.
This model was also used to model global brain state that fluctuates between two levels of excitability in anesthetized cortex~\cite{Curto2009}.
We use the following parameter values $a = -0.1$, $b = 0.01$, $c = 0.02$ and $I = 0.1$ to simulate 100 trajectories of 1000 steps with step size $0.5$ and Gaussian noise (std=$0.002$).
At this regime, unlike the ring attractor, the spontaneous dynamics is a periodic oscillation, and the trajectory follows a limit cycle.
The point process observations were also sampled via the observation model of the same parametric form as that of the ring attractor example.
We used 20 radial basis functions for dynamic model and 100 hidden units for recognition model. While training the model, the input was clamped to zero, and expect the model to learn the spontaneous oscillator.

\begin{figure}[!ht]
	\centering
	\includegraphics[width=\textwidth]{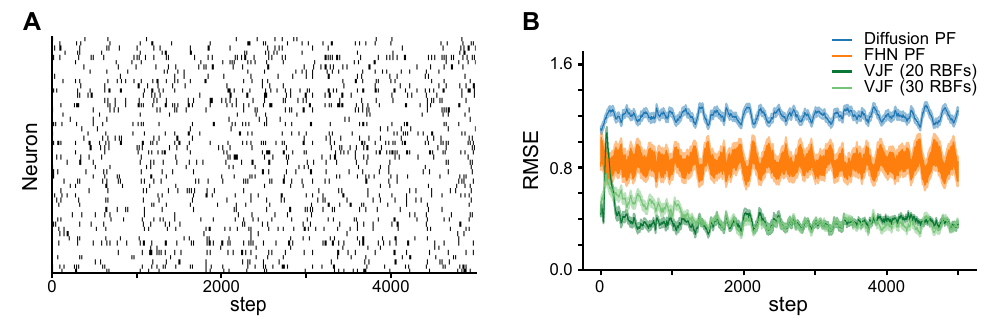}
	\caption[FitzHugh-Nagumo]{
		Nonlinear oscillator (FitzHugh-Nagumo) observation and state estimation.
		\textbf{(A)} The synthetic spike train.
		\textbf{(B)} RMSEs (the shades are s.e.m.) of state estimation on the observation in (A). We report the RMSE of estimated states among the proposed method VJF and two particle filters, one with diffusion dynamics and the other with the true FHN dynamics. Varying the number of RBFs did not substantially change the quality of the results.
	}
	\label{fig:pf}
\end{figure}

We compare the state estimation with the standard particles filtering (PF) which are powerful online methods theoretically capable of producing arbitrarily accurate filtering distribution.
We run two variants of the particle filter with different proposal distributions.
One used diffusion as the proposal, i.e. $\vx_t = \vx_{t-1} + \epsilon_t$ where $\vx$ is the vector of state variables $v$ and $w$, and the other, a.k.a. bootstrap particle filter~\cite{Gordon1993}, used the true dynamics in Eq.~\eqref{eq:FHN}.
We provided the true parameters for the observation model and noise term to PF which gives them an advantage.
Both particle filters and VJF were run on 50 realizations of 5000-step long observation series.
Figure~\ref{fig:pf} shows the root mean squared deviations (RMSE) (mean and standard error over 50 realizations). It is expected that the bootstrap particle filter outperformed the diffusion particle filter since the former utilized the true dynamics. One can see the state estimation by VJF improved as learning carrying on and eventually outperformed both particle filters. Note that VJF had to learn the parameters of likelihood, dynamic model and recognition model during the run.
We varied the number of RBFs (20 and 30) but the results are not substantially different.

\begin{figure}[!ht]
	\centering
	\includegraphics[width=\textwidth]{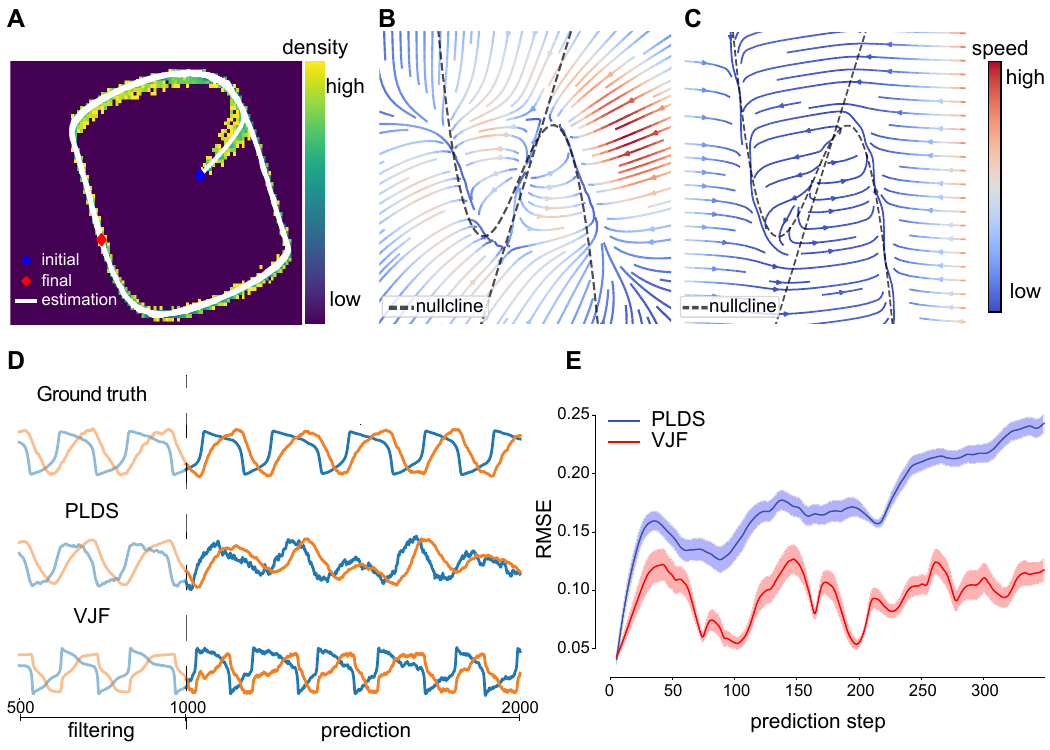}
	\caption[FitzHugh-Nagumo model]{
		Nonlinear oscillator (FitzHugh-Nagumo) dynamical system and prediction.
		\textbf{(A)} One inferred latent trajectory and the density of posterior means of all trajectories. Most of the inferred trajectory lie on the oscillation path.
		\textbf{(B)} Velocity field reconstructed by the inferred dynamical system.
		\textbf{(C)} Velocity field of the true dynamical system. The dashed lines are two nullclines of the true model on which the gradients are zero so as the velocity.
		\textbf{(D)} 1000-step prediction continuing the trajectory and sampled spike trains compared to ground truth from (A).
		\textbf{(E)} Mean (solid line) and standard error (shade) of root mean square error of prediction of 2000 trials. The prediction started at the same states for the true system and models.
		Note that PLDS fails to predict long term due to its linear dynamics assumption. A linear dynamical system without noise can only produce damped oscillations.
	}
	\label{fig:fhn}
\end{figure}

We also reconstructed the phase portrait (Fig.~\ref{fig:fhn}B) comparing to the truth (Fig.~\ref{fig:fhn}C). The two dashed lines are the theoretical nullclines of the true model on which the velocity of corresponding dimension is zero. The reconstructed field shows a low speed valley overlapping with the nullcline especially on the right half of the figure. At the intersection of the two nullclines there is an unstable fixed point. We can see the identified fixed point is close to the intersection. As most of the trajectories lie on the oscillation path (limit cycle) with merely few data points elsewhere, the inferred system shows the oscillation dynamics similar to the true system around the data region. The difference mostly happens in the region far from the trajectories because of the lack of data.

We run a long-term prediction using VJF without seeing the future data $\vy_{t+1:T}$ during these steps ($T=1000~\mathrm{steps} = 1~\text{sec}$) beginning at the final state of training data. We show the truth and prediction in figure~\ref{fig:fhn}D. The upper row is the true latent trajectory and corresponding observations. The lower row is the filtered trajectory and prediction by the proposed method. The light-colored parts are the 500 steps of inference before prediction and the solid-colored parts are 1000 steps truth and prediction. We also show the sample observations from the trained observation model during the prediction period.

One of the popular latent process modeling tools for point process observation that can make prediction is the Poisson Linear Dynamical System (PLDS)~\cite{Macke2011} which assumes latent linear dynamics.
We compared PLDS fit with EM on its long-term prediction on both the states and spike trains (Fig.~\ref{fig:fhn}). This demonstrates the nonlinear dynamical model outperforming the linear model even in the unfair online setting.

To compare to the methods with nonlinear dynamical models, we also run latent factor analysis via dynamical systems (LFADS)~\cite{Pandarinath2018} offline using the same data. LFADS implements its dynamical model with the gated recurrent unit (GRU)~\cite{Cho2014} that requires high dimensions. For this 2D system, we tried different GRU dimensionalities. We made minimal changes to its recommended setting including only the generator dimensionality, batch and no controller. The result shows that LFADS requires much higher dimension than the true system to capture the oscillation (Fig.~S1). (The figure of its inferred trajectories is shown in the supplement.) We report the fitted log-likelihood per time bin $-0.1274$, $-0.1272$ and $-0.1193$ for 2D, 20D and 50D GRU respectively. In comparison, the log-likelihood of the proposed approach is $-0.1142$ with a 2D dynamical model (higher the better).

\subsection{Fixed point attractor for decision-making}

\begin{figure}[!ht]
	\centering
	\includegraphics[width=\textwidth]{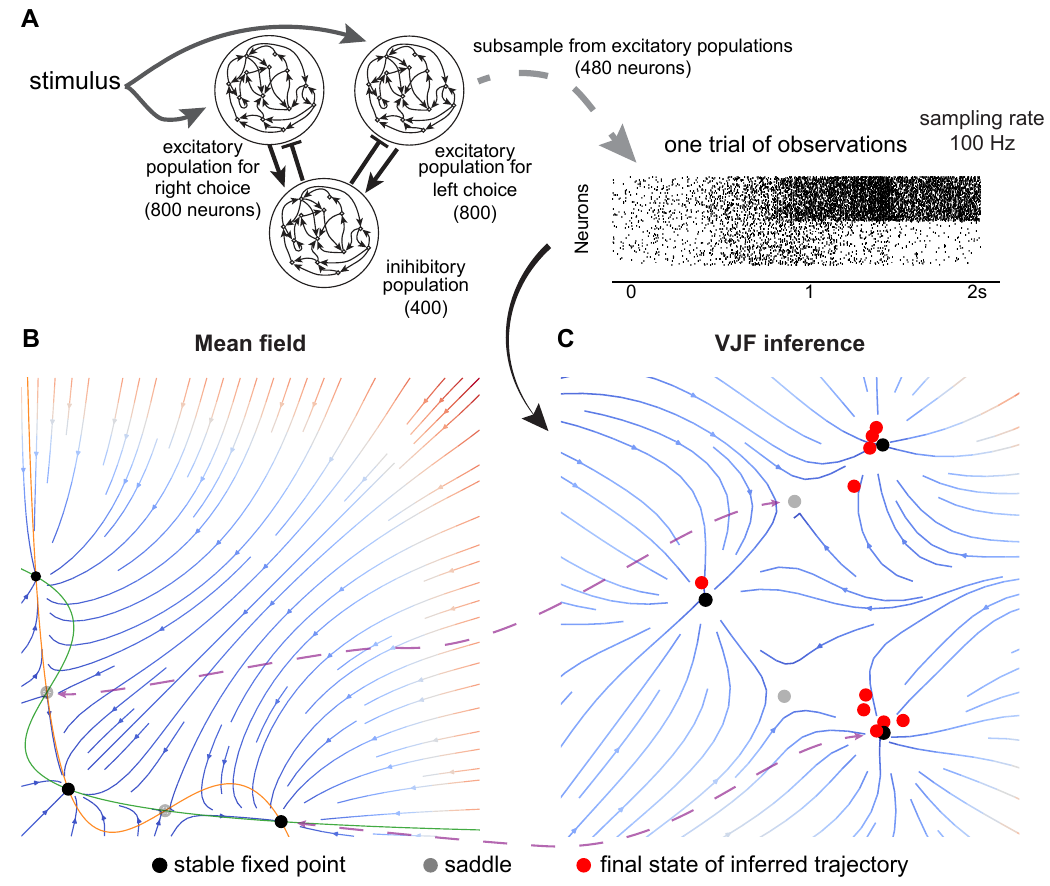}
	\caption[]{
		Fixed point attractor for decision-making.
		\textbf{(A)} Schematics of the neural network. There are two excitatory populations that are wired with slow recurrent excitation and feedback inhibition to produce attractor dynamics.
		The simulation was organized into decision-making trials.
		Each trial begins with a 0.5s period of spontaneous activity, and then the input is given to the two excitatory populations for 1.5s.
		We subsampled 480 selective neurons out of 1600 excitatory neurons from the simulation to be observed by our algorithm.
		\textbf{(B)} Mean field reduction of the network. Theoretical work has shown that the collective population dynamics can be reduced to 2 dimensions~\cite{Wong2006}.
		\textbf{(C)} VJF inferred dynamical model.
		The red dots are the inferred final states of zero-coherent trials.
		The black dots are fixed points (the solid are stable and the gray are unstable).
		Although the absolute arrangement is dissimilar, the topology and relation of the five identified fixed points show correspondence (indicated by purple lines).
	}\label{fig:wang}
\end{figure}

Perceptual decision-making paradigm is a well-established cognitive task where typically a low-dimensional decision variable needs to be integrated over time, and subjects are close to optimal in their performance.
To understand how the brain implements such neural computation, many competing theories have been proposed~\cite{Barak2013,Mante2013,Ganguli2008,Wang2002,Wong2006}.
We test our method on a simulated biophysically realistic cortical network model for a visual discrimination experiment~\cite{Wang2002}.
In the model, there are two excitatory subpopulations that are wired with slow recurrent excitation and feedback inhibition to produce attractor dynamics with two stable fixed points (Fig.~\ref{fig:wang}A).
Each fixed point represents the final perceptual decision, and the network dynamics amplify the difference between conflicting inputs and eventually generates a binary choice.

Note that, different from the former examples that use a linear-nonlinear map of latent states, the point process observations (spikes) of this experiment were directly sampled from the spiking neural network~\footnote{The detail of the spiking neural network can be found in~\cite{Wang2002} and the code can be found at~\url{https://github.com/xjwanglab/book/tree/master/wang2002}.}(1~ms binwidth) that was governed by its own high-dimensional intrinsic dynamics. It is filling the gap between fully specified state space models and real neuron populations.

We subsampled 480 selective neurons out of 1600 excitatory neurons from the simulation to be observed by our algorithm.
The simulated data is organized into decision-making trials where each trial lasts for 2 sec and with different strength of visual evidence, controlled by ``coherence''.
Our method with 20 radial basis functions learned the dynamics from 140 training trials (20 per coherence level $c$, $c=-1, -0.2, -0.1, 0, 0.1, 0.2, 1$).

Figure~\ref{fig:wang}C shows the velocity field at zero coherence stimulus as colored streamlines.
Note that our approach did not have prior knowledge of the network dynamics as the mean-field reduction~\cite{Wong2006} in Figure~\ref{fig:wang}B.
Although the absolute arrangement is dissimilar, the topology and relation of the five identified fixed points show correspondence with the mean-field reduction.
The inference was completely data-driven (partial observation of spike trains) while the mean-field method required knowing the true dynamical model of the network and careful approximation by \cite{Wong2006}.
We showed that our method can provide a qualitatively similar result to the theoretical work which reduces the dimensionality and complexity of the original network.

\subsection{Chaotic dynamics}

\begin{figure}[!ht]
	\centering
	\includegraphics[width=\textwidth]{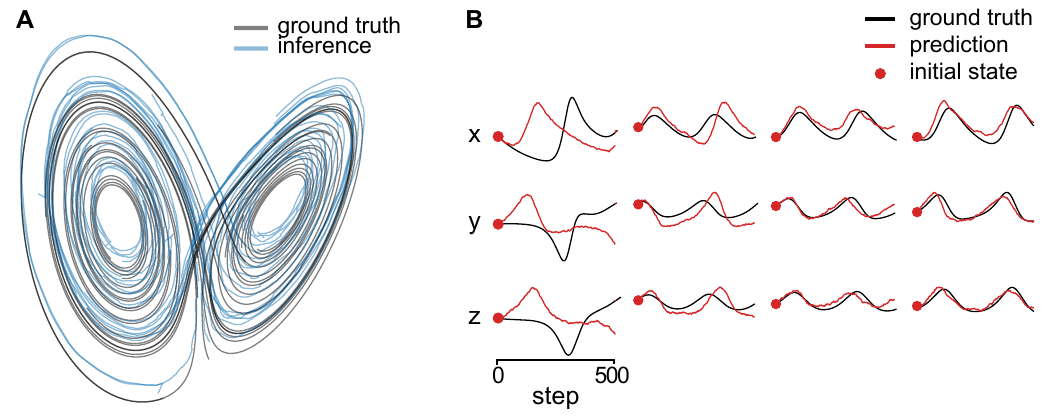}
	\caption[]{Lorenz attractor. \textbf{(A)} Estimated state trajectory (blue) and the ground truth (black) in 3D. \textbf{(B)} We predict 500 steps of future latent states (without knowing the respective observations) starting from 4 different initial states (red dots) using the inferred dynamical system. The red lines are the prediction and the black lines are the corresponding ground truth state.
	}
	\label{fig:lorenz}
\end{figure}

Chaotic dynamics (or edge-of-chaos) have been postulated to support asynchronous states in the cortex, and neural computation over time by generating rich temporal patterns~\cite{Maass2002,Laje2013}.
We consider the 3-dimensional standard Lorenz attractor as an example chaotic system to demonstrate the flexibility of our method.
We simulated 216 latent trajectories from:
\begin{equation}\label{eq:lorenz}
	\dot{x} = 10(y - x), \quad
	\dot{y} = x(28 - z) - y, \quad
	\dot{z} = x y - \frac{8}{3} z.
\end{equation}
The each coordinate of the initial states are on the uniform grid of 6 values in $[-50, 50]$ inclusively, of which the combination results in 216 unique states.
We discarded the first 500 transient steps of each trajectory and then use the following 1000 steps.
We generated 200-dimensional Gaussian observations driven by the trajectories.
Figure~\ref{fig:lorenz}A shows estimated latent trajectory and the ground truth. One can see that the estimation lies in a similar manifold.
In addition, we predicted 500 steps of future latent states without knowing the respective observations. Figure~\ref{fig:lorenz}B shows 4 predicted trajectories starting from different initial states. One can see that the inferred system could generate qualitatively similar trajectory at most initial states but not perfectly for the true system is chaotic.

\subsection{Nonstationary system}
\begin{figure}[!ht]
	\centering
	\includegraphics[width=0.7\textwidth]{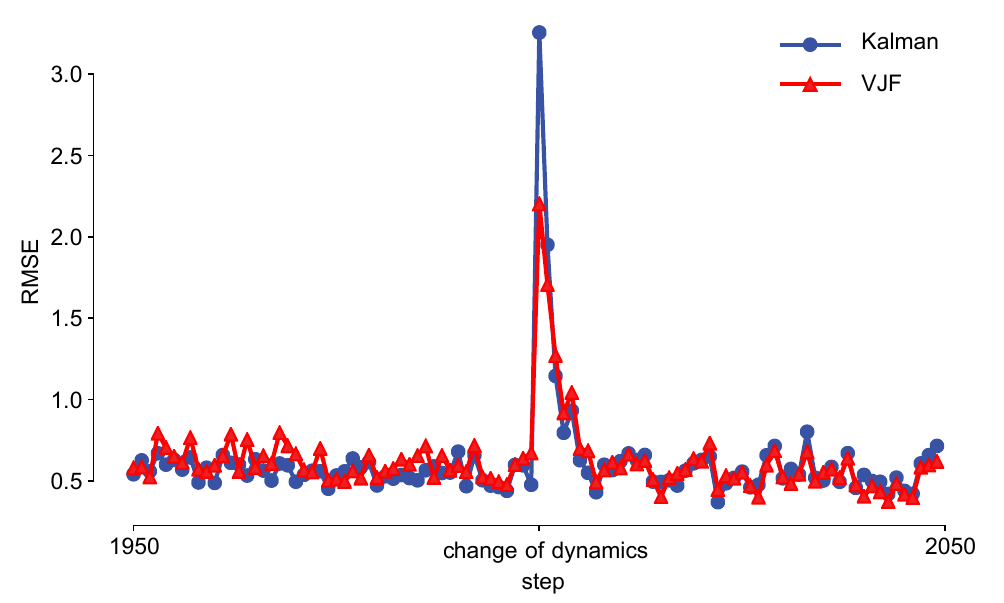}
	\caption[]{
		Prediction of nonstationary dynamical system. The colored curves (blue: EKF, red: VJF) are the mean RMSEs of one-step-ahead prediction of nonstationary system during online learning (50 trials). The linear system was changed and the state was perturbed at the 2000th step (center). The lines are average RMSEs. Both online algorithms quickly learned the change after a few steps.
	}\label{fig:nonstationary}
\end{figure}

Another feature of our method is that its state dynamics estimate never stops.
As a result, the algorithm is adaptive, and can potentially track slowly varying (nonstationary) latent dynamics.
To test this feature, we compared a dual EKF and the proposed approach on nonstationary linear dynamical system. A spiral-in linear system was suddenly changed from clockwise to counter-clockwise at the 2000th step and the latent state was perturbed (Fig.~\ref{fig:nonstationary}).
To adapt EKF, we used Gaussian observations that were generated through linear map from 2-dimensional state to 200-dimensional observation with additive noise ($\mathcal{N}(0, 0.5)$). To focus on the dynamics, we fixed all the parameters except the transition matrix for both methods, while our approach still has to learn the recognition model in addition.
Figure~\ref{fig:nonstationary} shows that our approach achieved better online performance as dual EKF in this experiment.

\section{Real Neurophysiological Application}

\begin{figure}[!ht]
	\centering
	\includegraphics[width=\textwidth]{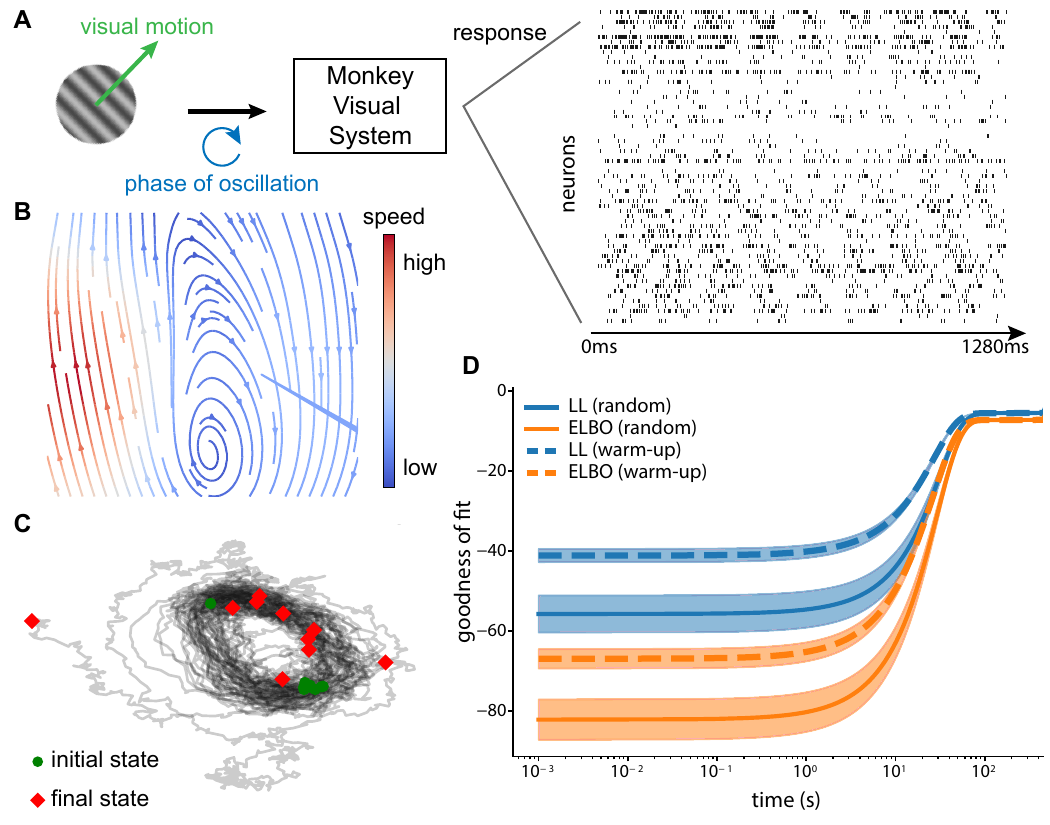}
	\caption[]{
		\textbf{(A)} Neurophysiological experiment. Drifting gratings were shown to the monkey (on the left). The neural spike trains (63 neurons, 1280~ms) from area V1 during the motion onset were recorded (on the right). Each row is one neuron and the binwidth is 1~ms. The phase of the oscillation forms a circular variable.
		\textbf{(B)} Phase portrait of the inferred dynamical system (arrows: direction, blue: low speed, red: high speed). The flow shows that the inferred system forms an oscillator.
		\textbf{(C)} Trajectories simulated from the inferred dynamical system. We simulated 10 state trajectories using the inferred system with random initial states (1000 steps each, black lines: trajectories, green circles: initial states, red diamonds: final states). The trajectories also confirm that the inferred system captured the oscillation underlying the data.
		\textbf{(D)} Convergence of the online method in terms of its goodness-of-fit.
		We calculated two goodness-of-fit measures (mean $\pm$ standard deviation, 10 repetitions), log-likelihood (LL) and ELBO for two strategies of initializing the observation model, warm-up and random initialization.
		Warm-up indicates that we initialized the observation model using dimensionality reduction methods before VJF; Random initialization indicates that the parameters of observation model were randomly drawn and learned completely by VJF.
	}\label{fig:v1}
\end{figure}

We applied the proposed method to a large scale recording to validate that it picks up meaningful dynamics. The dataset~\cite{Graf2011} consists of 148 simultaneously recorded single units from the primary cortex (V1) while directional drifting gratings were presented to an anesthetized monkey for around 1.3s per trial (Fig.~\ref{fig:v1}A). We used the spike trains from 63 well-tuned units. The spike times were binned with a 1ms window (max 1 spike per bin).
There is one continuous circular variables in the stimuli space: temporal phase of oscillation induced by the drifting gratings.

A partial warm-up helps with the training. We chose a good initialization for the observation model, specifically the loading matrix and bias. There are 72 motion directions in total, each repeated 50 trials.
We used the trials corresponding to 0 deg direction to initialize the observation model with dimensionality reduction methods such as variational latent Gaussian processes~\cite{Zhao2017}, and then trained VJF with a 2D dynamic model fully online on the trials corresponding to 180 deg direction that it had not seen before. Since we do not have long enough continuously-recorded trials, we concatenated the trials (equivalent to 500s) as if they were continuously recorded to mimic an online setting.
As expected, Figure~\ref{fig:v1}B and C shows the inferred dynamical system is able to implement the oscillation.
The two goodness of fit measures (log-likelihood and ELBO) in Figure~\ref{fig:v1}D shows that our method benefits from but does not necessarily require such a warm-up. The model with warm-up initialization has better starting goodness of fit than the random initialized model but the random initialized model eventually achieved similar goodness of fit with adequate amount of data.

\section{Discussion}

Neurotechnologies for recording the activity of large neural populations during meaningful behavior provide exciting opportunities for investigating the neural computations that underlie perception, cognition, and decision-making.
However, the datasets provided by these technologies currently require sophisticated offline analyses that slow down the scientific cycle of experiment, data analysis, hypothesis generation, and further experiment.
Moreover, in closed-loop neurophysiological setting, real-time adaptive algorithms are extremely valuable~\cite{Jordan2019b}.

To fulfill this demand, we proposed an online algorithm for recursive variational Bayesian inference that simultaneously performs system identification and state filtering under the framework of state space modeling, in hope that it can greatly impact neuroscience research and biomedical engineering.
There is no other method capable of all features, hence we compared several methods in different measures, often giving them the advantage.
We showed that our proposed method consistently outperforms the state-of-the-art methods.

Using the language of dynamical systems, we interpret the target system not via model parameters but via dynamical features: fixed points, limit cycles, strange attractors, bifurcations and so on.
In our current approach, this interpretation heavily relies on visual inspection of the qualitative nonlinear dynamical system features.
In contrast, most popular state space models assume linear dynamics~\cite{Ho1964,Macke2011,Katayama2005} which is appropriate for smoothing latent states, but not expressive enough to recover the underlying vector field.
Recently the Koopman theory that allows representation of general nonlinear dynamics as linear operators in infinite dimensional spaces~\cite{Koopman1931} has gained renewed interest in modeling nonlinear dynamics.
Although elegant in theory, in practice, however, the Koopman operators need to be truncated to a finite dimensional space with linear dynamics~\cite{Brunton2016}.
We note that the resulting linear models do not allow for topological features such as multiple isolated fixed points, nonlinear continuous attractors, stable limit cycles---features critical for non-trivial neural computation.

Our algorithm is highly flexible and general---it allows a wide range of observation models (likelihoods) and dynamic models, is computationally tractable, and produces interpretable visualizations of complex collective network dynamics.
Our key assumption is that the dynamics consists of a continuous and slow flow, which enable us to parameterize the velocity field directly.
This assumption reduces the complexity of the nonlinear function approximation, thus it is easy to identify the fixed/slow points.
Specifically we chose the radial basis function network to model the dynamics for our experiments, which regularizes and encourages the dynamics to occupy a finite phase volume around the origin.

Our method has a number of hyperparameters.
In the experiments, the differentiable hyperparameters were learnt via gradient descent while the selection of the other hyperparameters were made simple.
In general, our method was robust; Perturbing the number of RBFs did not produce qualitatively different results (Fig.~\ref{fig:pf}).
\cite{Liu2009} discussed growing radial basis function network adaptively which could be incorporated in our method to enable online tuning of the number of RBFs.
The depth and width of neural networks were chosen empirically to improve the interpretability of resulting dynamical systems, but tuning did not result in large changes in the results.

This work opens many avenues for future work.
One direction is to apply this model to large-scale neural recording from a behaving animal.
We hope that further development would enable on-the-fly analysis of high-dimensional neural spike train during electrophysiological experiments.
Clinically, a nonlinear state space model provides a basis for nonlinear feedback control as a potential treatment for neurological diseases that arise from diseased dynamical states.

\bibliographystyle{hunsrt}
\bibliography{ref}

\begin{thebibliography}{10}

\bibitem{Haykin1998}
S.~Haykin and J.~Principe.
\newblock Making sense of a complex world [chaotic events modeling].
\newblock {\em {IEEE} Signal Processing Magazine}, 15(3):66--81, May 1998.

\bibitem{Ko2009}
J.~Ko and D.~Fox.
\newblock {GP}-{BayesFilters}: Bayesian filtering using gaussian process
  prediction and observation models.
\newblock {\em Autonomous Robots}, 27(1):75--90, 5 2009.

\bibitem{Mattos2016}
C.~L.~C. Mattos, Z.~Dai, A.~Damianou, et~al.
\newblock Recurrent gaussian processes.
\newblock {\em International Conference on Learning Representations (ICLR)},
  2016.

\bibitem{Roweis2001}
S.~Roweis and Z.~Ghahramani.
\newblock {\em Learning nonlinear dynamical systems using the
  expectation-maximization algorithm}, pages 175--220.
\newblock John Wiley \& Sons, Inc, 2001.

\bibitem{Mante2013}
V.~Mante, D.~Sussillo, K.~V. Shenoy, and W.~T. Newsome.
\newblock Context-dependent computation by recurrent dynamics in prefrontal
  cortex.
\newblock {\em Nature}, 503(7474):78--84, November 2013.

\bibitem{Sussillo2013}
D.~Sussillo and O.~Barak.
\newblock Opening the black box: Low-dimensional dynamics in high-dimensional
  recurrent neural networks.
\newblock {\em Neural Computation}, 25(3):626--649, March 2013.

\bibitem{Frigola2014}
R.~Frigola, Y.~Chen, and C.~E. Rasmussen.
\newblock Variational gaussian process state-space models.
\newblock In {\em Proceedings of the 27th International Conference on Neural
  Information Processing Systems - Volume 2}, pages 3680--3688, Montreal,
  Canada, 2014.

\bibitem{Zhao2017}
Y.~Zhao and I.~M. Park.
\newblock Variational latent {G}aussian process for recovering single-trial
  dynamics from population spike trains.
\newblock {\em Neural Computation}, 29(5), May 2017.

\bibitem{Zhao2016}
Y.~Zhao and I.~M. Park.
\newblock Interpretable nonlinear dynamic modeling of neural trajectories.
\newblock In {\em Advances in Neural Information Processing Systems (NIPS)},
  2016.

\bibitem{Russo2018}
A.~A. Russo, S.~R. Bittner, S.~M. Perkins, et~al.
\newblock Motor cortex embeds muscle-like commands in an untangled population
  response.
\newblock {\em Neuron}, 97(4):953--966, feb 2018.

\bibitem{Ho1964}
Y.~Ho and R.~Lee.
\newblock A {B}ayesian approach to problems in stochastic estimation and
  control.
\newblock {\em IEEE Transactions on Automatic Control}, 9(4):333--339, October
  1964.

\bibitem{Sarkka2013}
S.~S\"arkk\"a.
\newblock {\em Bayesian filtering and smoothing}.
\newblock Cambridge University Press, 2013.

\bibitem{Haykin2001}
S.~S. Haykin.
\newblock {\em {K}alman filtering and neural networks}.
\newblock Wiley, 2001.

\bibitem{Ghahramani1999}
Z.~Ghahramani and S.~T. Roweis.
\newblock Learning nonlinear dynamical systems using an {EM} algorithm.
\newblock In M.~J. Kearns, S.~A. Solla, and D.~A. Cohn, editors, {\em Advances
  in Neural Information Processing Systems 11}, pages 431--437. MIT Press,
  1999.

\bibitem{Valpola2002}
H.~Valpola and J.~Karhunen.
\newblock An unsupervised ensemble learning method for nonlinear dynamic
  {State-Space} models.
\newblock {\em Neural Computation}, 14(11):2647--2692, November 2002.

\bibitem{Turner2010}
R.~Turner, M.~Deisenroth, and C.~Rasmussen.
\newblock State-space inference and learning with gaussian processes.
\newblock In Y.~W. Teh and M.~Titterington, editors, {\em Proceedings of the
  Thirteenth International Conference on Artificial Intelligence and
  Statistics}, volume~9 of {\em Proceedings of Machine Learning Research},
  pages 868--875, Chia Laguna Resort, Sardinia, Italy, 13--15 May 2010. PMLR.

\bibitem{Golub2013}
M.~D. Golub, S.~M. Chase, and B.~M. Yu.
\newblock Learning an internal dynamics model from control demonstration.
\newblock {\em JMLR workshop and conference proceedings}, pages 606--614, 2013.

\bibitem{Archer2015}
E.~Archer, I.~M. Park, L.~Buesing, J.~Cunningham, and L.~Paninski.
\newblock Black box variational inference for state space models.
\newblock {\em ArXiv e-prints}, November 2015.

\bibitem{Krishnan2015}
R.~G. Krishnan, U.~Shalit, and D.~Sontag.
\newblock Deep {K}alman filters.
\newblock {\em arXiv}, abs/1511.05121, November 2015.

\bibitem{Johnson2016}
M.~Johnson, D.~K. Duvenaud, A.~Wiltschko, R.~P. Adams, and S.~R. Datta.
\newblock Composing graphical models with neural networks for structured
  representations and fast inference.
\newblock In D.~D. Lee, M.~Sugiyama, U.~V. Luxburg, I.~Guyon, and R.~Garnett,
  editors, {\em Advances in Neural Information Processing Systems 29}, pages
  2946--2954. Curran Associates, Inc., 2016.

\bibitem{Krishnan2017}
R.~G. Krishnan, U.~Shalit, and D.~Sontag.
\newblock Structured inference networks for nonlinear state space models.
\newblock {\em arXiv}, abs/1511.05121, 2016.

\bibitem{Karl2017}
M.~Karl, M.~Soelch, J.~Bayer, and P.~van~der Smagt.
\newblock Deep variational {B}ayes filters: Unsupervised learning of state
  space models from raw data.
\newblock In {\em 5th International Conference on Learning Representations},
  2017.

\bibitem{Watter2015}
M.~Watter, J.~Springenberg, J.~Boedecker, and M.~Riedmiller.
\newblock Embed to control: A locally linear latent dynamics model for control
  from raw images.
\newblock In C.~Cortes, N.~D. Lawrence, D.~D. Lee, M.~Sugiyama, and R.~Garnett,
  editors, {\em Advances in Neural Information Processing Systems 28}, pages
  2746--2754. Curran Associates, Inc., 2015.

\bibitem{Broderick2013}
T.~Broderick, N.~Boyd, A.~Wibisono, A.~C. Wilson, and M.~I. Jordan.
\newblock Streaming variational bayes.
\newblock In C.~J.~C. Burges, L.~Bottou, M.~Welling, Z.~Ghahramani, and K.~Q.
  Weinberger, editors, {\em Advances in Neural Information Processing Systems
  26}, pages 1727--1735. Curran Associates, Inc., 2013.

\bibitem{Wan2000}
E.~A. Wan and R.~Van Der~Merwe.
\newblock The unscented {K}alman filter for nonlinear estimation.
\newblock In {\em Proceedings of the IEEE 2000 Adaptive Systems for Signal
  Processing, Communications, and Control Symposium (Cat. No.00EX373)}, pages
  153--158, Lake Louise, Alta., Canada, August 2000. IEEE.

\bibitem{Wan2001}
E.~A. Wan and A.~T. Nelson.
\newblock {\em Dual extended {K}alman filter methods}, pages 123--173.
\newblock John Wiley \& Sons, Inc, 2001.

\bibitem{Cover1991}
T.~M. Cover and J.~A. Thomas.
\newblock {\em Elements of Information Theory}.
\newblock Wiley-Interscience, August 1991.

\bibitem{Hinton1995}
G.~Hinton, P.~Dayan, B.~Frey, and R.~Neal.
\newblock The "wake-sleep" algorithm for unsupervised neural networks.
\newblock {\em Science}, 268(5214):1158--1161, May 1995.

\bibitem{Hastie2009}
T.~Hastie, R.~Tibshirani, and J.~Friedman.
\newblock {\em The Elements of Statistical Learning}.
\newblock Springer-Verlag GmbH, 2009.

\bibitem{Sussillo2016}
D.~Sussillo, R.~Jozefowicz, L.~F. Abbott, and C.~Pandarinath.
\newblock {LFADS} - latent factor analysis via dynamical systems.
\newblock {\em arXiv}, abs/1608.06315, August 2016.

\bibitem{Rezende2014}
D.~J. Rezende, S.~Mohamed, and D.~Wierstra.
\newblock Stochastic backpropagation and approximate inference in deep
  generative models.
\newblock In {\em International Conference on Machine Learning}, May 2014.

\bibitem{Kingma2014}
D.~P. Kingma and M.~Welling.
\newblock Auto-{Encoding} {Variational} {Bayes}.
\newblock {\em arXiv:1312.6114 [cs, stat]}, May 2014.
\newblock arXiv: 1312.6114.

\bibitem{Kingma2014a}
D.~P. Kingma and J.~Ba.
\newblock Adam: {A} method for stochastic optimization.
\newblock {\em CoRR}, abs/1412.6980, 2014.

\bibitem{Newman2015}
J.~P. Newman, M.-f. Fong, D.~C. Millard, et~al.
\newblock Optogenetic feedback control of neural activity.
\newblock {\em eLife}, 2015.

\bibitem{ElHady2016}
A.~El~Hady.
\newblock {\em Closed Loop Neuroscience}.
\newblock Academic Press, London, United Kingdom, 2016.

\bibitem{Hocker2019a}
D.~Hocker and I.~M. Park.
\newblock Myopic control of neural dynamics.
\newblock {\em PLOS Computational Biology}, 2019.

\bibitem{Macke2011}
J.~H. Macke, L.~Buesing, J.~P. Cunningham, et~al.
\newblock Empirical models of spiking in neural populations.
\newblock In J.~Shawe-Taylor, R.~S. Zemel, P.~L. Bartlett, F.~Pereira, and
  K.~Q. Weinberger, editors, {\em Advances in Neural Information Processing
  Systems 24}, pages 1350--1358. Curran Associates, Inc., 2011.

\bibitem{Hopfield1982}
J.~J. Hopfield.
\newblock Neural networks and physical systems with emergent collective
  computational abilities.
\newblock {\em Proceedings of the National Academy of Sciences},
  79(8):2554--2558, April 1982.

\bibitem{Dayan2001}
P.~Dayan and L.~F. Abbott.
\newblock {\em Theoretical neuroscience: computational and mathematical
  modeling of neural systems}.
\newblock Massachusetts Institute of Technology Press, 2001.

\bibitem{Barak2013}
O.~Barak, D.~Sussillo, R.~Romo, M.~Tsodyks, and L.~F. Abbott.
\newblock From fixed points to chaos: three models of delayed discrimination.
\newblock {\em Progress in neurobiology}, 103:214--222, April 2013.

\bibitem{Wang2002}
X.-J. Wang.
\newblock Probabilistic decision making by slow reverberation in cortical
  circuits.
\newblock {\em Neuron}, 36(5):955--968, December 2002.

\bibitem{Nassar2018b}
J.~Nassar, S.~W. Linderman, M.~Bugallo, and I.~M. Park.
\newblock Tree-structured recurrent switching linear dynamical systems for
  multi-scale modeling.
\newblock In {\em International Conference on Learning Representations (ICLR)},
  November 2019.

\bibitem{Pillow2008}
J.~W. Pillow, J.~Shlens, L.~Paninski, et~al.
\newblock Spatio-temporal correlations and visual signalling in a complete
  neuronal population.
\newblock {\em Nature}, 454(7207):995--999, July 2008.

\bibitem{Truccolo2005}
W.~Truccolo, U.~T. Eden, M.~R. Fellows, J.~P. Donoghue, and E.~N. Brown.
\newblock A point process framework for relating neural spiking activity to
  spiking history, neural ensemble, and extrinsic covariate effects.
\newblock {\em Journal of Neurophysiology}, 93(2):1074--1089, February 2005.

\bibitem{Strogatz2000}
S.~H. Strogatz.
\newblock {\em Nonlinear Dynamics and Chaos}.
\newblock Studies in nonlinearity. The Perseus Books Group, January 2000.

\bibitem{Peyrache2015}
A.~Peyrache, M.~M. Lacroix, P.~C. Petersen, and G.~Buzsaki.
\newblock Internally organized mechanisms of the head direction sense.
\newblock {\em Nature Neuroscience}, 18(4):569--575, March 2015.

\bibitem{Izhikevich2007}
E.~M. Izhikevich.
\newblock {\em Dynamical systems in neuroscience : the geometry of excitability
  and bursting}.
\newblock Computational neuroscience. MIT Press, 2007.

\bibitem{Curto2009}
C.~Curto, S.~Sakata, S.~Marguet, V.~Itskov, and K.~D. Harris.
\newblock A simple model of cortical dynamics explains variability and state
  dependence of sensory responses in {Urethane-Anesthetized} auditory cortex.
\newblock {\em The Journal of Neuroscience}, 29(34):10600--10612, August 2009.

\bibitem{Gordon1993}
N.~J. Gordon, D.~J. Salmond, and A.~F.~M. Smith.
\newblock Novel approach to nonlinear/non-gaussian bayesian state estimation.
\newblock {\em {IEE} Proceedings F Radar and Signal Processing}, 140(2):107,
  1993.

\bibitem{Pandarinath2018}
C.~Pandarinath, D.~J. O'Shea, J.~Collins, et~al.
\newblock Inferring single-trial neural population dynamics using sequential
  auto-encoders.
\newblock {\em Nature Methods}, 15(10):805--815, sep 2018.

\bibitem{Cho2014}
K.~Cho, B.~van Merrienboer, D.~Bahdanau, and Y.~Bengio.
\newblock On the properties of neural machine translation: Encoder-decoder
  approaches.
\newblock 2014.

\bibitem{Ganguli2008}
S.~Ganguli, J.~W. Bisley, J.~D. Roitman, et~al.
\newblock One-dimensional dynamics of attention and decision making in {LIP}.
\newblock {\em Neuron}, 58(1):15--25, April 2008.

\bibitem{Wong2006}
K.-F. Wong and X.-J. Wang.
\newblock A recurrent network mechanism of time integration in perceptual
  decisions.
\newblock {\em The Journal of Neuroscience}, 26(4):1314--1328, January 2006.

\bibitem{Maass2002}
W.~Maass, T.~Natschl\"ager, and H.~Markram.
\newblock Real-time computing without stable states: A new framework for neural
  computation based on perturbations.
\newblock {\em Neural Computation}, 14:2531--2560, 2002.

\bibitem{Laje2013}
R.~Laje and D.~V. Buonomano.
\newblock Robust timing and motor patterns by taming chaos in recurrent neural
  networks.
\newblock {\em Nature Neuroscience}, 16(7):925--933, May 2013.

\bibitem{Graf2011}
A.~B.~A. Graf, A.~Kohn, M.~Jazayeri, and J.~A. Movshon.
\newblock Decoding the activity of neuronal populations in macaque primary
  visual cortex.
\newblock {\em Nature Neuroscience}, (2):239--245, January 2011.

\bibitem{Jordan2019b}
I.~D. Jordan and I.~M. Park.
\newblock Birhythmic analog circuit maze: A nonlinear neurostimulation testbed.
\newblock {\em Entropy}, 22(5):537, May 2020.

\bibitem{Katayama2005}
T.~Katayama.
\newblock {\em Subspace methods for system identification}.
\newblock Springer, 2005.

\bibitem{Koopman1931}
B.~O. Koopman.
\newblock Hamiltonian systems and transformation in hilbert space.
\newblock {\em Proceedings of the National Academy of Sciences of the United
  States of America}, 17(5):315--318, May 1931.

\bibitem{Brunton2016}
B.~W. Brunton, L.~A. Johnson, J.~G. Ojemann, and J.~N. Kutz.
\newblock Extracting spatial{\textendash}temporal coherent patterns in
  large-scale neural recordings using dynamic mode decomposition.
\newblock {\em Journal of Neuroscience Methods}, 258:1--15, jan 2016.

\bibitem{Liu2009}
W.~Liu, I.~Park, and J.~C. Principe.
\newblock An information theoretic approach of designing sparse kernel adaptive
  filters.
\newblock {\em {IEEE} Transactions on Neural Networks}, 20(12):1950--1961, dec
  2009.

\end{thebibliography}

\end{document}